\documentclass[twoside,11pt,english,utf8ttf]{article}

\usepackage{jmlr2e}
\usepackage{amsmath}
\usepackage{enumitem}
\usepackage{hyperref}

\setlistdepth{9}
\newlist{myEnumerate}{enumerate}{9}
\setlist[myEnumerate,1]{label=\arabic*.}
\setlist[myEnumerate,2]{label=(\alph*)}
\setlist[myEnumerate,3]{label=\Alph*.}
\setlist[myEnumerate,4]{label=\roman*.}
\setlist[myEnumerate,5]{label=(\alph*)}
\setlist[myEnumerate,6]{label=(\arabic*)}
\setlist[myEnumerate,7]{label=(\Roman*)}
\setlist[myEnumerate,8]{label=(\Alph*)}
\setlist[myEnumerate,9]{label=(\roman*)}

\hypersetup{colorlinks=true, linkcolor=blue}

\jmlrheading{1}{2000}{1-48}{4/00}{10/00}{Abhinav Maurya}

\ShortHeadings{Detecting Events in Text Streams}{Abhinav Maurya}
\firstpageno{1}

\begin{document}

\title{Spatial Semantic Scan: Jointly Detecting Subtle Events and their Spatial Footprint}

\author{\name Abhinav Maurya \email ahmaurya@cmu.edu \\
\addr Heinz College \\
Carnegie Mellon University \\
Pittsburgh, PA -- 15213 \\
 \\
}

\editor{Leslie Pack Kaelbling}

\maketitle

\begin{abstract}
Many methods have been proposed for detecting emerging events in text streams using topic modeling. However, these methods have shortcomings that make them unsuitable for rapid detection of locally emerging events on massive text streams. We describe Spatially Compact Semantic Scan (SCSS) that has been developed specifically to overcome the shortcomings of current methods in detecting new spatially compact events in text streams. SCSS employs alternating optimization between using semantic scan (\cite{liu2011detecting}) to estimate contrastive foreground topics in documents, and discovering spatial neighborhoods (\cite{shao2011generalized}) with high occurrence of documents containing the foreground topics. We evaluate our method on Emergency Department chief complaints dataset (ED dataset) to verify the effectiveness of our method in detecting real-world disease outbreaks from free-text ED chief complaint data.
\end{abstract}

\begin{keywords}
Event Detection, Latent Dirichlet Allocation, Topic Modeling
\end{keywords}

\newpage
\tableofcontents
\newpage

\section{Introduction}
\label{sec:introduction}

Text streams are ubiquitous in data processing and knowledge discovery workflows. Their analysis and summarization is difficult because of their unstructured nature, the sparsity of the canonical bag-of-words representation, the massive scale of web-scale text streams like Twitter and Yelp Reviews, and the noise present due to word variations from mispellings, dialects, and slang.

Topic modeling is a mixed-membership model used to summarize a corpus of text documents from a set of latent topics, where each topic is a sparse distribution on words. However, traditional topic modeling methods like Latent Dirichlet Allocation (LDA) are too slow for analyzing web-scale text streams, and also assume that there is no concept drift in the topics being learned over time. Variations like \emph{Online LDA} (\cite{hoffman2010online}), \emph{Dynamic Topic Models} (\cite{blei2006dynamic}), \emph{Topics over Time} (\cite{wang2006topics}), and \emph{Non-parametric Topics over Time} (\cite{dubey2013nonparametric}) relax the assumption that there is no concept drift in the learned topics with time, but make strong assumptions about the evolution of topics with time.


In this paper, we propose \emph{Spatially Compact Semantic Scan (SCSS)} which was developed to overcome these shortcomings in the scalable detection of spatially localized emerging topics in text streams. 

In the \emph{Background} section (\ref{sec:background}), we introduce important terminology used in the rest of the paper, describe Latent Dirichlet Allocation (LDA), and collapsed Gibbs sampling used in LDA inference. In the \emph{Related Work} section (\ref{sec:related_work}), we present a literature survey and compare SCSS to related previous work on event detection in text streams. In the \emph{Methodology} section (\ref{sec:methodology}), we motivate and describe \emph{Spatially Compact Semantic Scan (SCSS)}. In the \emph{Results} section (\ref{sec:results}), we present results comparing SCSS to state-of-the-art methods described in the \emph{Related Work} section. We discuss the results and possible avenues for future work in the \emph{Discussions} section (\ref{sec:discussions}), and conclude the report in the \emph{Conclusions} section (\ref{sec:conclusions}).

\section{Background}
\label{sec:background}

\subsection{Terminology}

Here, we introduce some terminology that we will encounter often through this report:-

\begin{enumerate}
	\item \textbf{Latent Dirichlet Allocation} (\cite{blei2003latent}): A Bayesian mixed-membership topic model which treats each text document as a mixture of various topics i.e. multinomial distributions over words. Described in section (\ref{sec:lda-generative-process}).
	\item \textbf{Semantic Scan (SS)}: A enhancement of the LDA topic model to detect emerging topics in text corpora. Described in detail in section (\ref{sec:ss-generative-process}).
	\item \textbf{Spatially Compact Semantic Scan (SCSS)}: A further improvement over Semantic Scan to detect emerging topics from spatio-temporal text corpora such that the emerging topic occurs in documents are spatially located close to each other. Described in detail in section (\ref{sec:scss-generative-process}).
	\item \textbf{Background Documents}: To detect emerging topics, we assume that a portion of our corpus is composed of documents where the emerging topic does not occur. These documents are referred to as background documents, and the portion of the corpus is referred to as background corpus.
	\item \textbf{Foreground Documents}: The documents which may contain the emerging topics are called foreground documents, and the appropriate portion of the corpus is called foreground corpus. The division of the entire corpus into foreground and background corpora is designated by the user of the method. The user needs to specify the dividing timestamp such that the documents that were collected before the timestamp are designated background documents and the documents that were collected after the timestamp are designated foreground documents. For the ED dataset, we chose all documents from 2003 as the background documents and all documents from 2004 as the foreground documents. 
	\item \textbf{Background Topics}: Also called old topics or static topics. These are topics that are considered to have generated the background documents through the LDA generative process.
	\item \textbf{Foreground Topics}: Also called new topics or emerging topics. These topics alongwith the background topics are considered to have generated the foreground documents through the LDA generative process.
\end{enumerate}

\subsection{Latent Dirichlet Allocation (LDA)}
\label{sec:lda-generative-process}

Here, we briefly describe the Latent Dirichlet Allocation (LDA) topic model (\cite{blei2003latent}) since it forms the foundation of SCSS. LDA assumes that documents are generated as a mixture of topics, where topics themselves are distributions over words. The model also has an intuitive polyhedral interpretation: the documents reside on a low-dimensional topic simplex embedded in the high-dimensional word simplex. Parameter inference on the LDA topic model therefore aims at dimensionality reduction using a small number of latent topics so as to best explain the observed documents using these latent topics. While this assumption is not necessarily true in practice,it is a very useful assumption that helps in denoising and discovering frequent distinctive word co-occurrences from text corpora.

\begin{figure}[ht]
\vskip 0.2in
\begin{center}
\centerline{\includegraphics[trim=60mm 90mm 30mm 80mm, clip=true, width=0.7\columnwidth]{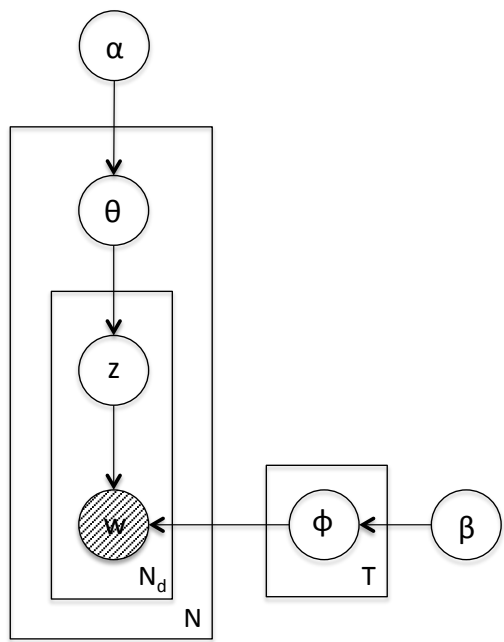}}
\caption{Plate Diagram of the LDA Topic Model}
\label{fig:lda-plate-model}
\end{center}
\vskip -0.2in
\end{figure} 

We refer the reader to figure (\ref{fig:lda-plate-model}) which describes the generative model for LDA. There are $T$ topics given by $\phi = \{\phi_i: i=1,..,T\}$, where each $\phi_i$ is a distribution over the $N_u$ unique words in the dictionary. Since $\phi_i$ is the parameter of a multinomial distribution over words, it is naturally assumed to be generated from a Dirichlet prior with parameter $\beta$. There are $N$ documents in total. The multinomial distribution over the $T$ topics in the $i^{th}$ document is parameterized by $\theta_i$, which is generated from a Dirichlet distribution with parameter $\alpha$. To generate the $j^{th}$ word of the $i^{th}$ document, we first sample the topic at that position $z_{ij}$ from $\theta_i$, and then use the sampled topic to sample the actual word $w_{ij}$ from a multinomial distribution with parameter $\phi_{z_{ij}}$.

The plate diagram shows plates for the $T$ topics, $N$ documents, and $N_d$ words within each documents, indicating that these templates need to be repeated the corresponding number of times to obtain the full probabilistic graphical model over which inference of the unknown parameters will be performed. The observed words $w_{ij}$ in the $j^{th}$ position of the $i^{th}$ document are indicated by darkened circles in the plate diagram, while the unobserved variables to be inferred are indicated by empty circles. Exact inference in LDA is intractable in general, and the posterior over the unobserved parameters is obtained using (collapsed) Gibbs sampling or variational inference. Recent research has found the LDA model to be identifiable under assumptions of separability i.e. each topic has an anchor word that only occurs in that topic with positive probability, and hence only appears in documents which have that topic in their generating mixture of topics (\cite{arora2012learning}). Further work has found that the model is identifiable under weaker assumptions (\cite{bansal2014provable}).

\subsection{Collapsed Gibbs Sampling for LDA Inference}
\label{sec:lda-collapsed-gibbs}

The most common sampling-based approach to LDA inference is collapsed Gibbs sampling (\cite{griffiths2004finding}). In collapsed Gibbs sampling, we maintain the following variables during the inference:-

\begin{enumerate}
	\item $\bf{Z}$ $ = \{z_{ij}\}$: topic assignments in all documents indexed by $i$ and word positions within each document indexed by $j$.
	\item $n_{ik}$: the number of times topic $k$ is assigned to words in document $i$. Therefore, $n_{ik} = \sum_{j} [z_{ij} == k]$. Here, $[condition]$ is the indicator function that is 1 when the $condition$ is true and 0 when the $condition$ is false. 
	\item $n_{kw}$: the number of times topic $k$ is assigned to word $w$ in the entire corpus $C$. Therefore, $n_{kw} = \sum_{i,j} [z_{ij} == k \text{ and } w_{ij} == w]$ i.e. we count all word positions in the corpus where the actual word is $w$ and the topic assignment is $k$.
	\item $n_k$: the number of times topic $k$ is assigned to any word in the entire corpus $C$. Therefore, $n_k = \sum_w n_{kw}$.
\end{enumerate}

We note that $n_{ik}$, $n_{kw}$, and $n_k$ can be calculated given $\bf{Z}$ as explained for each one of them. However, these statistics are stored because having access to them at each step of the Gibbs sampling makes the process much faster.

Let ${\bf Z}_{-ij}$ denote a particular instance of topic assignments, excluding the assignment at the $j^{th}$ position of the $i^{th}$ document. Let $n^{(-ij)}_{ik}$, $n^{(-ij)}_{kw}$, and $n^{(-ij)}_k$ be the $n_{ik}$, $n_{kw}$, and $n_k$ aggregate statistics calculated also without considering the topic assignment at the $j^{th}$ position of the $i^{th}$ document. These can be easily calculated from $n_{ik}$, $n_{kw}$, and $n_k$ by subtracting the contribution to these counts resulting from the topic assignment $z_{ij}$ at $j^{th}$ position of the $i^{th}$ document.

Collapsed Gibbs sampling proceeds through all words in the corpus sequentially. At each word position $w_{ij}$ in document $i$, it calculates the multinomial topic probability $P(z_{ij})$ conditioned on the observed corpus $D$ and all other topic assignments in the corpus ${\bf Z}_{-ij}$. The calculation is governed by the following formula:

\begin{equation}
P(z_{ij} = k \mid {\bf Z}_{-ij}, {\bf D}) \propto (n^{(-ij)}_{ik}+\alpha_k) \left(
\frac{n^{(-ij)}_{kw} + \beta_w}{ n^{(-ij)}_{k} + \sum_w \beta_w} \right)
\label{eq:collapsed_gibbs_lda}
\end{equation}

Gibbs sampling then samples a topic from the topic multinomial distribution $P(z_{ij} = k \mid {\bf Z}_{-ij}, {\bf D})$ and updates the statistics $n_{ik}$, $n_{kw}$, and $n_k$ by adding the count contribution from the newly sampled topic $z_{ij}$ at word position $w_{ij}$. The algorithm then moves to the next word position and repeats the topic multinomial calculation, topic sampling, and statistics updates. This process of sequentially going through the words of the corpus and sampling $z_{ij}$ is known to eventually converge to sampling from the stationary distribution of the LDA topic model after an intial burn-in period typical of MCMC sampling methods.

\section{Related Work}
\label{sec:related_work}

In this section, we briefly describe some related papers and discuss some of their shortcomings in detecting spatially compact emerging topics in text streams.

\subsection{Efficient topic model inference on streaming document collections}
\label{rel:yao2009efficient}

The \emph{Gibbs2} and \emph{Gibbs3} sampling-based inference methods described in (\cite{yao2009efficient}) are very similar to the Semantic Scan setting we describe below, and perhaps the closest work in literature to which we can compare our method. Gibbs2, Gibbs3, and Semantic Scan begin by learning topic assignments for words in the background documents and then begin inference on the foreground documents. All three methods also hold the topic assignments for words in background documents fixed, while performing sampling for topic assignments in the foreground documents. However, a key distinction is that our method allows additional new topics to be assigned to words in foreground documents, while Gibbs2 and Gibbs3 do not. We will see that allowing new topics to be learned entirely from foreground documents leads to precise topics that characterize emerging events in the text stream well. In fact, setting the number of new topics in Semantic Scan to 0 gives us Gibbs3. This is because the background topics are not allowed to change once they have been learned in both Semantic Scan and Gibbs3. Thus, Semantic Scan generalizes Gibbs3 for the purpose of emerging event detection.

\begin{figure}[ht]
\vskip 0.2in
\begin{center}
\centerline{\includegraphics[width=0.6\columnwidth]{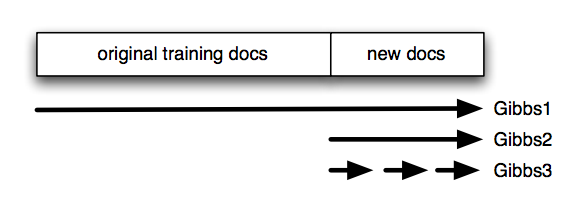}}
\caption{Schematic diagram showing the distinction between Gibbs1, Gibbs2, and Gibbs3. Note that SCSS is similar to Gibbs2 where we learn new topics from the foreground documents in a batch fashion. From (\cite{yao2009efficient})}
\label{fig:gibbs123-schematic-diagram}
\end{center}
\vskip -0.2in
\end{figure}

\subsection{Labeled LDA}
\label{rel:ramage2009labeled}

Labeled LDA (\cite{ramage2009labeled}) is another paper which closely resembles our experimental setup. It is a supervised method in which different partitions of the corpus are constrained to contain different sets of topics. This is helpful in a multi-labeled text corpus where each text document can possibly be assigned multiple labels by human labelers to explicitly indicate the topics it contains. Each label is then associated with a topic and a document is assumed to contain only the topics corresponding to its labels during the Gibbs sampling. In our setup, we can associate the background documents with a subset of topics assigned to the foreground documents and the corrrespondence of our method with labeled LDA really becomes clear. However, this simplistic solution ignores the fact that the number of foreground documents available for emerging event detection may be orders of magnitude smaller than the number of background documents collected over years or even decades. Applying labeled LDA naively would mean performing Gibbs sampling on the entire corpus of background and foreground documents every time we receive a batch of new documents. In the SCSS method, we need to perform Gibbs sampling only on the foreground documents which can be much more computationally efficient than labeled LDA.

\subsection{Topics over time}
\label{rel:wang2006topics}

\emph{Topics over Time: A Non-Markov Continuous-Time Model of Topical Trends
} (\cite{wang2006topics}) presents a graphical model which relaxes the assumption of Markovian evolution of the natural parameters of the topic model. Instead, each topic is associated with a continuous beta distribution over timestamps normalized to the interval $[0,1]$. The topics remain static over time, however, the occurrence of topics in the corpus varies with time. However, the assumption that the number of topics is constant over time and that only the topic parameters evolve smoothly with time is still present in this model.

Non-paramteric topics over time (\cite{dubey2013nonparametric}) is a variation of the algorithm that allows the number of topics to be determined from the corpus. However, the topics are still constrained to evolve smoothly over time.

\subsection{Online LDA}
\label{rel:hoffman2010online}

Online LDA (\cite{hoffman2010online}) employs online variational Bayes inference to determine the posterior distribution over the latent variables of the topic model. The algorithm is based on online stochastic optimization and is shown to provide equivalently good topics in lesser time compared to batch variational Bayes algorithm. The algorithm requires a learning rate $k \in (0.5,1]$ for convergence. This parameters specifies the rate at which the old parameters are forgotten. Thus, there is an assumption of parameter smoothness which can delay the detection of suddenly emerging topics in a text stream.

\subsection{Online NMF}
\label{rel:cao2007latent}

\emph{Latent Factor Detection and Tracking with Online Non-Negative Matrix Factorization} (\cite{cao2007latent}) suggests using currently learnt topics (as a proxy to past documents) along with the new documents to learn the new set of topics. This approach is similar to a variant of semantic scan where the background topics are used in the initialization step of the MCMC procedure to find foreground topics, but are not held fixed through the inference of foreground topics. The drawback is that the foreground topics found using this method might include the background topics in their span but may not help us precisely find the emerging foreground topics, since the detected topics might be a mixture of both the old and the new latent factors.

\subsection{Kernel Topic Models}
\label{rel:hennig2012kernel}

\emph{Kernel Topic Models} (\cite{hennig2012kernel}) is a topic model that can incorporate spatial, temporal, hierarchical, and social metadata about text documents in the topic model by assuming that a document is represented by real-valued features that are generated by real-valued functions sampled from a Gaussian process prior, and passing these features through a softmax function to obtain the document-topic proportions that lie on the probability simplex. It is a adaptation of Gaussian process latent variable model (GPLVM) (\cite{lawrence2003gaussian}) where the document-topic proportions are obtained in a manner similar to GPLVM but the topics are sampled from a Dirichlet prior as in LDA. Kernel Topic Models make no distinction between background and foreground documents and might be unable to detect spatially localized emerging events in a small number of foreground documents compared to a large background corpus. Like many other models, Kernel Topic Models also cannot be applied to event detection since it does not have the ability to detect when no event is emerging in the foreground documents without significant modification or additions to the algorithm.

\subsection{Adaptive Topic Models}
\label{rel:alsumait2008line}

(\cite{alsumait2008line}) propose an online version of LDA for topic detection and tracking. However, the method makes a strong assumption about the evolution of topics: the parameter of the Dirichlet prior generating a topic is a linear combination of the topic vector from the previous $\delta$ iterations of the algorithm. The smoothness and strict form imposed on the evolution of topics will not allow the method to detect rapidly emerging topics or subtle spatially localized topics hidden in the stream. In addition, the assumption is that the number of topics is constant over time and only the topic parameters evolve smoothly with time. There is no reason to believe that this is true, since the addition of a new topic does not mean that an old topic has disappeared from the corpus.

\subsection{Dynamic Topic Models}
\label{rel:blei2006dynamic}

\begin{figure}[ht]
\vskip 0.2in
\begin{center}
\centerline{\includegraphics[width=0.6\columnwidth]{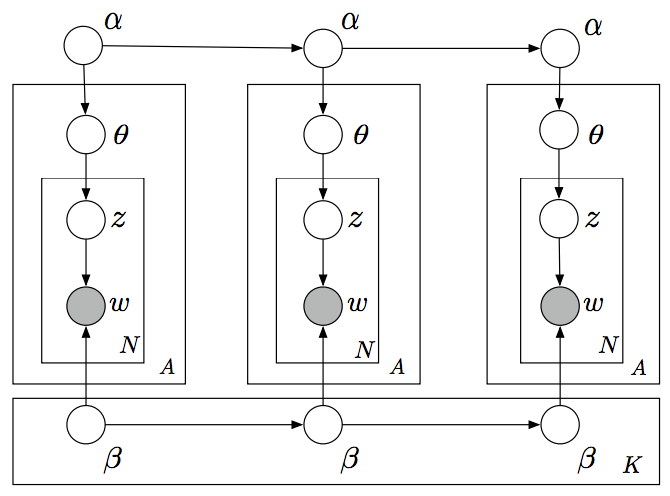}}
\caption{Graphical representation of a dynamic topic model (for three time slices). From (\cite{blei2006dynamic})}
\label{fig:dtm-plate-model}
\end{center}
\vskip -0.2in
\end{figure}

\emph{Dynamic Topic Models} (\cite{blei2006dynamic}) extends the LDA model by allowing the natural multinomial parameters of LDA to evolve over consecutive time slices. This is the standard Markovian assumption of state space models. The model is best illustrated using the plate diagram shown in figure (\ref{fig:dtm-plate-model}) which shows how DTM extends the LDA topic model shown in figure (\ref{fig:lda-plate-model}) and clearly illustrates the Markovian evolution of parameters in the topic model.

\subsection{A Latent Variable Model for Geographic Lexical Variation}
\label{rel:eisenstein2010latent}

(\cite{eisenstein2010latent}) propose a hierarchical LDA model consisting of a set of pure topics which suffer variations with region to form regional topics that finally generate the documents. The model assumes a fixed number of regions, and that pure topics exist in the form of regional variants in every region. A region is modeled using a bivariate Gaussian distribution which assumes that each region has a center where its regional topics are concentrated and the effect of the regional topics decays away from this center. While consideriing a bivariate Gaussian distribution for modeling the location of documents in SCSS is possible, it implies that the effect of a topic decays away from the epicenter. This might not be true in the case of steady state of a disease outbreak where all documents in the affected spatial neighborhood may be equally likely to contain the emerging topic.

\subsection{Discovering Geographical Topics in the Twitter Stream}
\label{rel:hong2012discovering}

(\cite{hong2012discovering}) propse another topic model for modeling text documents annotated with geospatial coordinates. As in (\cite{eisenstein2010latent}), (\cite{hong2012discovering}) also model the locations of documents as drawn from a bivariate Gaussian distribution. However, the goal is to model geographically localized topics where each topic is dominant in one region, rather than discover regional variants of pure topics as is the case with (\cite{eisenstein2010latent}). Both (\cite{eisenstein2010latent}) and (\cite{hong2012discovering}) deal only with the spatial aspect of topic models and do not address detection of an emerging spatially localized topic.



\subsection{Topic Posterior Contraction Analysis}
\label{rel:posterior_contraction}

Recent work (\cite{tang2014understanding}) suggests a theoretical justification as to why typical topic models do not work well on short documents like tweets. This justifies the additional novel contributions we need to incorporate in topic modeling to improve the outcome of topic modeling on a spatio-temporal corpus of short text documents.

\section{Methodology}
\label{sec:methodology}

\begin{table}[t]
\caption{Notation used in this paper}
\label{tab:notation}
\vskip 0.15in
\begin{center}
\begin{small}
\begin{sc}
\begin{tabular}{ll}
\hline
Symbol & Explanation \\
\hline
$N_b$      		& number of background documents \\
$N_f$      		& number of foreground documents \\
$N = N_b + N_f$	& total number of documents \\
$D_b$      		& corpus consisting of background documents \\
$D_f$      		& corpus consisting of foreground documents \\
$D$      		& composite corpus consisting of documents from $D_b$ and $D_f$\\
$D_{bi}$      	& $i^{th}$ background document in $D_b$ \\
$D_{fi}$      	& $i^{th}$ foreground document in $D_f$ \\
$D_i$      		& $i^{th}$ document in $D$ \\
$w_{bi}$      	& words in $i^{th}$ background document in $D_b$ \\
$w_{fi}$      	& words in $i^{th}$ foreground document in $D_f$ \\
$w_i$      		& words in $i^{th}$ document in $D$ \\
$T_b$      		& number of background topics \\
$T_f$      		& number of foreground topics \\
$T = T_b + T_f$	& total number of topics \\
$N_d$      		& number of words in a single document \\
$N_u$      		& number of unique words in the dictionary \\
$\eta$      	& hyperparameter for distribution of $\gamma$ \\
$\gamma$    	& severity hyperparameter for $\delta$ \\
$s_c$       	& center of spatial region \\
$n$         	& size of spatial region \\
$S_{cn}$    	& set of nodes in spatial region \\
$p$         	& sparsity parameter \\
$S$         	& a subset of $S_{cn}$ \\
$\delta$    	& variable capturing if doc has new topic \\
$\alpha_b$    	& dirichlet hyperparameter for mixture of old topics \\
$\alpha$   		& dirichlet hyperparameter for mixture of all topics\\
$\phi_b$      	& background topics from $1..T_b$\\
$\phi_f$      	& foreground topics from $1..T_f$\\
$\phi$      	& all topics from $1..(T_b+T_f)$\\
$\beta_b$     	& dirichlet hyperparameter for generating background topics\\
$\beta_f$     	& dirichlet hyperparameter for generating foreground topics\\
$\theta$    	& multinomial parameter for document-specific topic mixture \\
$z$         	& sampled topic per word position \\
$w$         	& sampled word at each position\\
\hline
\end{tabular}
\end{sc}
\end{small}
\end{center}
\vskip -0.1in
\end{table}

\begin{figure}
\begin{center}
    \begin{tabular}{c  c}
    \includegraphics[trim=70mm 90mm 30mm 90mm, clip=true, width=0.4\columnwidth]{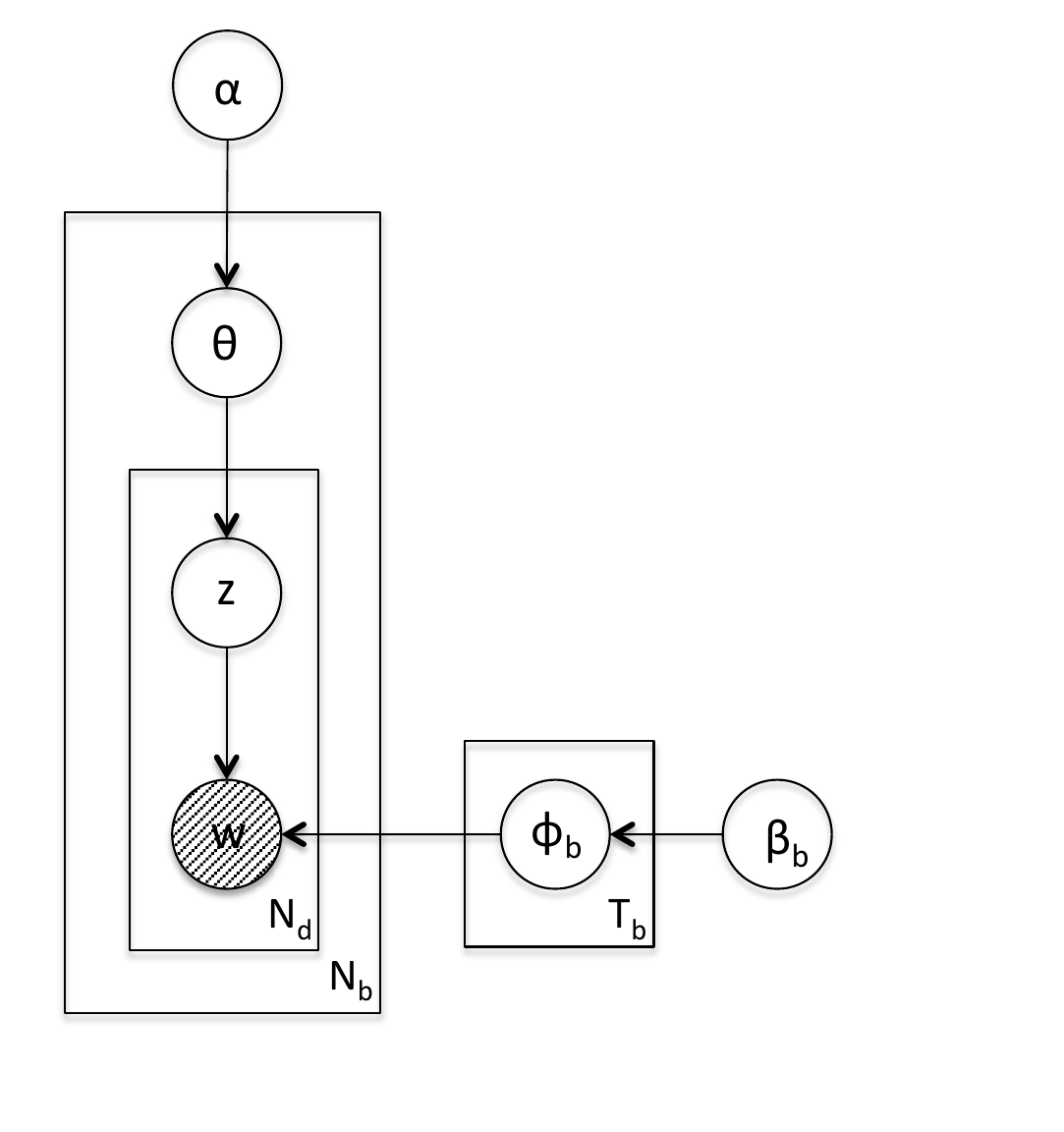} & \includegraphics[trim=70mm 90mm 30mm 90mm, clip=true, width=0.4\columnwidth]{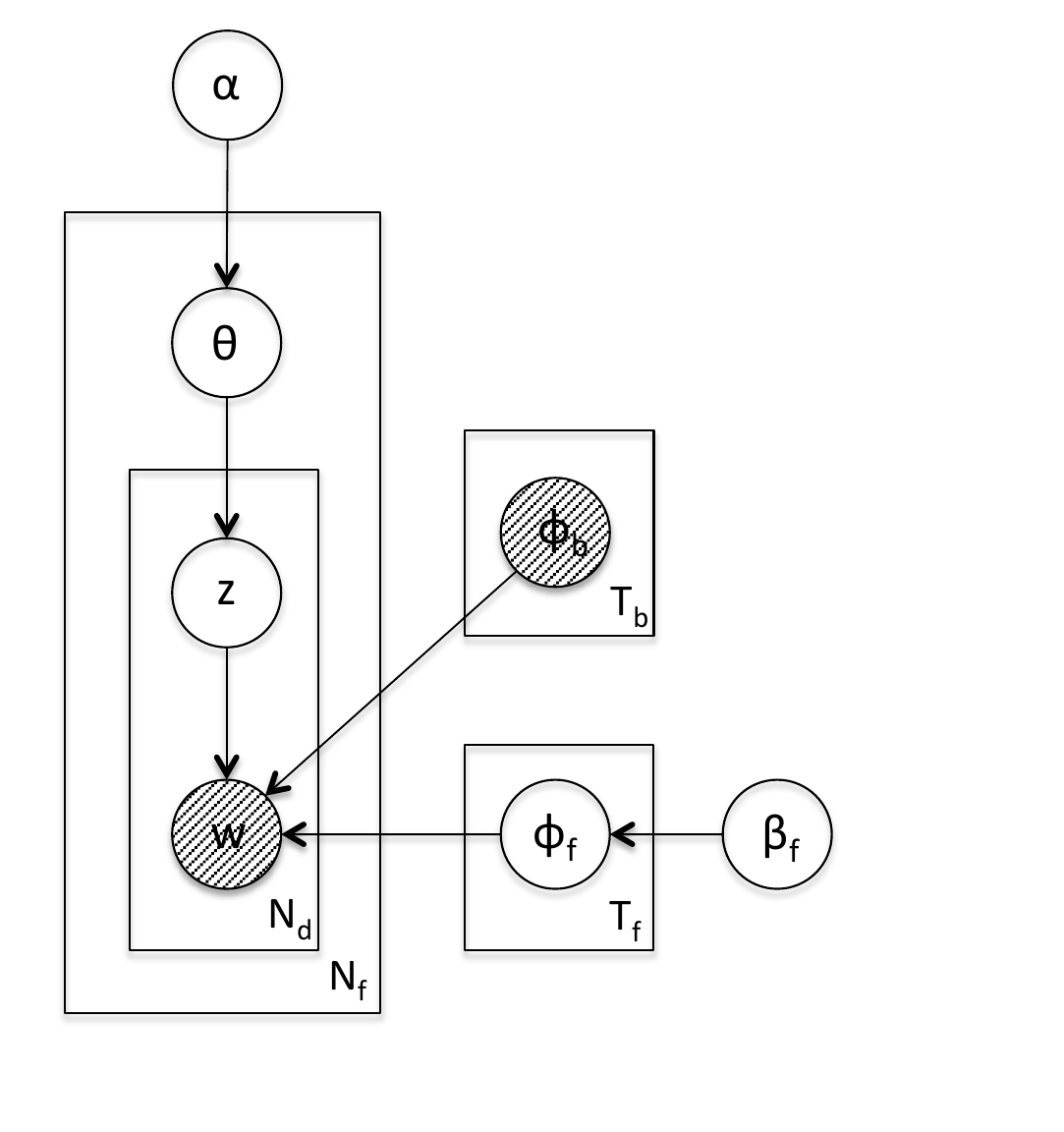} \\
	(a) Semantic Scan First Phase & (b) Semantic Scan Second Phase \\
    \end{tabular}
\end{center}
\caption{Plate Diagram of Spatially Compact Semantic Scan}
\label{fig:ss-plate-model}
\end{figure}

\begin{figure}[ht]
\vskip 0.2in
\begin{center}
\centerline{\includegraphics[trim=40mm 60mm 20mm 70mm, clip=true, width=0.6\columnwidth]{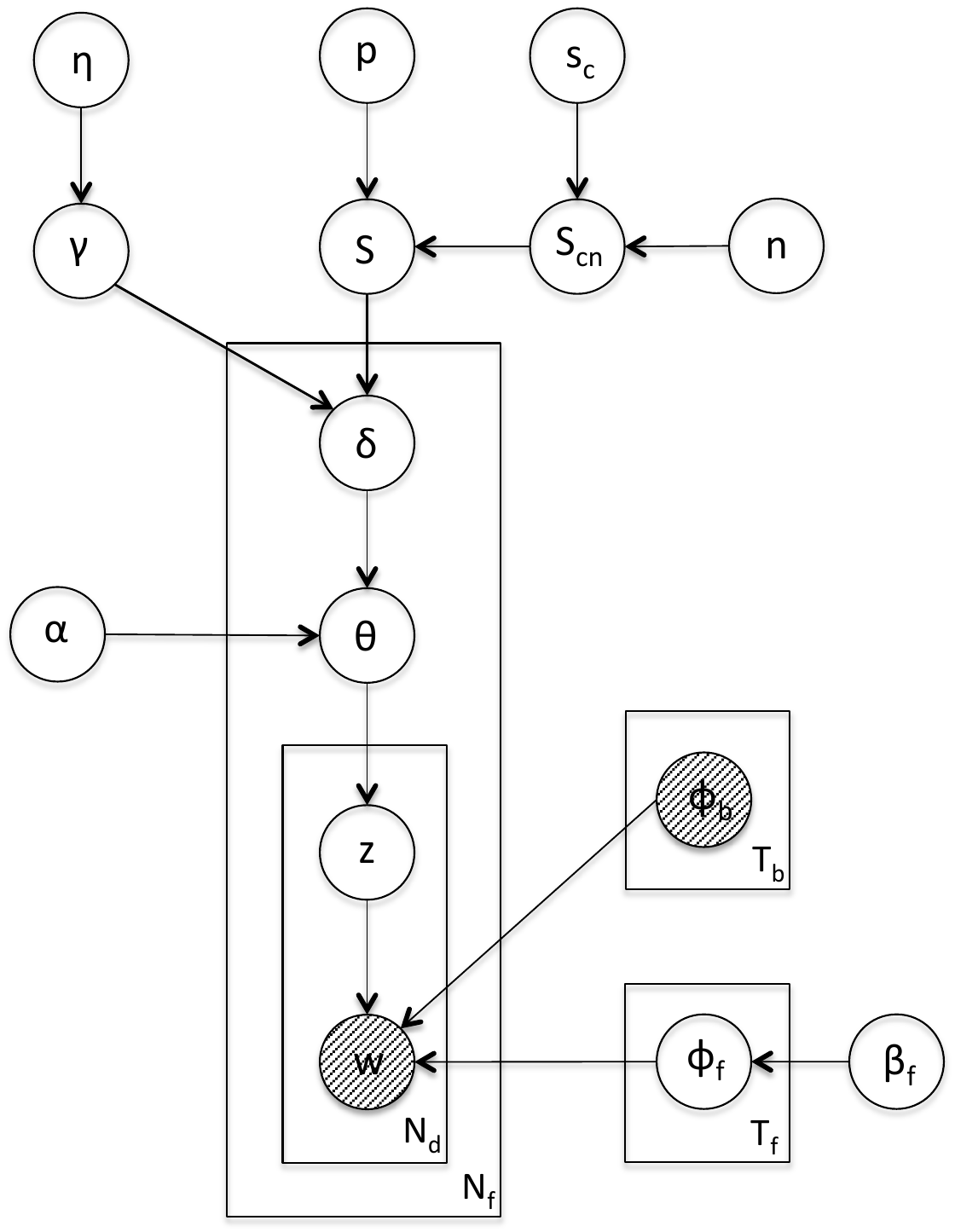}}
\caption{Plate Diagram of Spatially Compact Semantic Scan.}
\label{fig:scss-plate-model}
\end{center}
\vskip -0.2in
\end{figure} 

In our method, we aim to capture spatial coherence of the emerging topic i.e. we want to detect documents which contain the emerging topic and are spatially close to each other. This spatial proximity can be interpreted in a broad sense. In the simplest case, it can mean that documents that contain the new topic are actually generated at geospatial locations close to each other. In a more sophisticated case, we can consider that documents which contain the new topic are closer in other ways. For example, tweets generated by users that follow each other on Twitter can be considered to be close to each other in the heterogeneous network structure of the Twitter social graph. This could be useful in detecting topics that spread virally on a social network. For example, a stock market crash or undue market volatality could be a source of discussion and tweets among traders in New York, London, and Hong Kong who follow each other. Although these cities are geographically distributed, we believe that market-related tweets emerging from these cities will have similarity on other measures derived from the underlying network structure.

\subsection{Semantic Scan}
\label{sec:ss-generative-process}

Semantic Scan (SS) \cite{liu2011detecting,murray2013semantic} is a method which learns temporally emerging events from a text stream. The method learns a set of background topics from background documents which do not contain any emerging event of interest. When a new set of documents comes in, semantic scan learns a new set of topics for these documents while contrasting them with already learnt background topics so as not to relearn old topics. We describe SS in detail here, because our method Spatially Compact Semantic Scan (SCSS) builds on SS to incorporate spatial cohesion of documents containing the emerging topic as well.

The original semantic scan learns the set of emerging topics without taking spatial information into account, then performs a spatial scan to identify emerging spatial cluster of documents assigned to these topics. In contrast, SCSS coherently integrates a spatial model of the affected locations with the emerging topic model of semantic scan, thus enabling it to learnmore precisely focused, spatially localized topics which improve overall detection performance.

The Semantic Scan (SS) model for detecting emerging topics is illustrated using the plate diagram in figure (\ref{fig:ss-plate-model}). The notation used in the model is explained in table (\ref{tab:notation}). The Bayesian generative model can be outlined as follows:-

\begin{enumerate}
	\item \textbf{Generate background documents comprising $C_b$}
	\begin{enumerate}
		\item Choose a $T_b$-dimensional Dirichlet hyperparameter $\alpha_b$ for generating document-specific topic distributions, where $T_b$ is the number of background topics.
		\item Choose a $W$-dimensional Dirichlet hyperparameter $\beta_b$ for generating background topics, where $W$ is the number of words in the vocabulary.
		\item Sample $T_b$ background topics from $Dir(\beta_b)$ together denoted by $\phi_b$; each topic $\phi_{bi}$ is a $W$-dimensional multinomial distribution over words.
		\item For each of the $i=1,..,D_b$ background documents:
		\begin{enumerate}
			\item Sample a $T_b$-dimensional multinomial distribution $\theta_i$ over background topics from $Dir(\alpha_b)$.
			\item For each of the $j=1,..,N_i$ words in the document:
			\begin{enumerate}
				\item Sample a topic $z_{ij}$ for the word position $j$ from $Mult(\theta_i)$. $z_{ij}$ is one of the background topics $\phi_b$.
				\item Sample a word $w_{ij}$ for the word position $j$ from $Mult(\phi_{bz_{ij}})$.
			\end{enumerate}
		\end{enumerate}
	\end{enumerate}
	\item \textbf{Generate foreground documents comprising $C_f$}
	\begin{enumerate}
		\item Choose a $T$-dimensional Dirichlet hyperparameter $\alpha$ for generating document-specific topic distributions, where $T=T_b+T_f$ is the total number of topics including background and foreground topics.
		\item Choose a $W$-dimensional Dirichlet hyperparameter $\beta_f$ for generating foreground topics.
		\item Sample $T_f$ foreground topics from $Dir(\beta_f)$ together denoted by $\phi_f$; each topic $\phi_{fi}$ is a $W$-dimensional multinomial distribution over words.
		\item Denote the set of all topics as $\phi$ which includes topics from $\phi_b$ indexed from $1,..,T_b$ and topics from $\phi_f$ indexed from $T_b+1,..,T_b+T_f$
		\item For each of the $i=1,..,D_f$ foreground documents:
		\begin{enumerate}
			\item Sample a $T$-dimensional multinomial distribution $\theta_i$ over all topics from $Dir(\alpha)$.
			\item For each of the $j=1,..,N_i$ words in the document
			\begin{enumerate}
				\item Sample a topic $z_{ij}$ for the word position $j$ from $Mult(\theta_i)$. $z_{ij}$ is one of the topics in $\phi$.
				\item Sample a word $w_{ij}$ for the word position $j$ from $Mult(\phi_{z_{ij}})$.
			\end{enumerate}
		\end{enumerate}
	\end{enumerate}
\end{enumerate}

\subsection{Inference for Semantic Scan}
\label{sec:ss-inference}

The inference procedure consists of two phases as shown in figure (\ref{fig:ss-plate-model}). In the first phase, we learn a set of background LDA topics $\phi_{bi},~i=1,..,T_b$ using collapsed Gibbs sampling on a set of background documents. In the second phase, we keep the first $T_b$ topics fixed and learn new topics $\phi_{bi},~i=T_b+1,..,T_b+T_f$ while allowing the document-topic distributions $\theta_i$ for the foreground documents $C_f$ to change during the Gibbs sampling procedure. For details of the LDA collapsed Gibbs sampling procedure, we refer the reader to section (\ref{sec:background}) and (\cite{griffiths2004finding}). If the number of background topics $T_b$ is sufficient and we have learnt the background topics well, then fixing the background topics during the detection of the new topic propels the new topic to capture emerging trends in the text stream. This is because the span of fixed background topics explains words in the documents that are produced by the background data-generating process and are irrelevant to the emerging topic.

The plate diagram in figure (\ref{fig:ss-plate-model}.b) shows the model on which inference is performed in the second phase. The model shows that the words of the documents as well as the $T_b$ topics learned from the first phase are observed variables in the second phase, and we perform collapsed Gibbs sampling inference to learn the emerging topics $\phi_f$.

After the topic modeling step, SS (\cite{murray2013semantic,liu2011detecting}) assigns each document to one of the topics using an EM-like approach, and performs circular expectation-based poission spatial scan in order to detect a circular neighborhood of zipcodes that are affected by the outbreak.

\subsection{Spatially Compact Semantic Scan}
\label{sec:scss-generative-process}

In order to ensure that the emerging topic is also \emph{spatially regularized} to occur in spatially nearby documents, we place a hierarchical prior over the spatial regions whose documents can be affected by the emergence of the new topic. The proposed topic model (which incorporates SS as a building block) is illustrated using the plate diagram in figure (\ref{fig:scss-plate-model}). The notation used in the plate diagram is described in table (\ref{tab:notation}). The document generation process is as follows: We first select a subset of zipcodes where the documents will contain a new topic. To do this, we select a center of the spatial region $s_c$ and a neighborhood size $n$. The set of all zipcodes that are the $n$ nearest neighbors of $s_c$ form a circular neighborhood $S_{cn}$. To construct an arbitrarily shaped spatial region $S$ from $S_{cn}$, we choose a sparsity parameter $p$ and sample the zipcodes from $S_{cn}$ with probability $p$. This gives us a set of zipcodes $S$. Documents from zipcodes in $S$ may contain the foreground topics. The $i^{th}$ zipcode is associated with severity $\gamma_i$ sampled from a Beta distribution parameterized by $\eta$, which indicates the proposrtion of documents at that zipcode that will contain foreground topics.

For $j^{th}$ document located at the $i^{th}$ zipcode, we sample a document-specific $\delta_j$ as the output of a Bernoulli experiment indicating if the document should contain the new topic using the severity of the emerging topic indicated by $\gamma_i$. Documents outside $S$ are generated using the old topics and its distribution characteristics. For documents outside $S$, we set their $\delta$ to 0. 

If $\delta_j$ is 1, we sample the distribution over topics $\theta_j$ using the hyperparameter $\alpha$ which indicates a distribution over all topics including the new ones. If $\delta_j$ is 0, we use $\alpha_b$ to sample the distribution over old topics only since the document is not supposed to contain the new topic. It is possible to constrain the new hyperparameter $\alpha$ for topic distributions using the old hyperparameter $\alpha_b$. However, in our case, both $\alpha_b$ and $\alpha$ are uniform symmetric priors.

Once we have sampled the multinomial parameters $\theta_j$, we can sample a topic for each word position in the document. This sampled topic can then be used to index into the set of topic vectors to get the parameters of a multinomial distribution over words. Using the topic chosen for the word position, we now can sample a word from the topic. This completes the generation process for the foreground documents in the text corpus with emerging spatially localized topics. The entire Bayesian generative model can be outlined as follows:

\begin{myEnumerate}
	\item \textbf{Generate background documents comprising $C_b$}
	\begin{myEnumerate}
		\item Choose a $T_b$-dimensional Dirichlet hyperparameter $\alpha_b$ for generating document-specific topic distributions.
		\item Choose a $W$-dimensional Dirichlet hyperparameter $\beta_b$ for generating background topics.
		\item Sample $T_b$ background topics from $Dir(\beta_b)$ together denoted by $\phi_b$; each topic $\phi_{bi}$ is a $W$-dimensional multinomial distribution over words.
		\item For each of the $i=1,..,D_b$ background documents:
		\begin{myEnumerate}
			\item Sample a timestamp unformly from the set of possible background timestamps $K_b$.
			\item Sample a zipcode uniformly from the set of zipcodes considered to generate the documents $Z$.
			\item Sample a $T_b$-dimensional multinomial distribution $\theta_i$ over background topics from $Dir(\alpha_b)$.
			\item For each of the $j=1,..,N$ words in the document:
			\begin{myEnumerate}
				\item Sample a topic $z_{ij}$ for the word position $j$ from $Mult(\theta_i)$. $z_{ij}$ is one of the background topics $\phi_b$.
				\item Sample a word $w_{ij}$ for the word position $j$ from $Mult(\phi_{bz_{ij}})$.
			\end{myEnumerate}
		\end{myEnumerate}
	\end{myEnumerate}
	\item \textbf{Generate foreground documents comprising $C_f$}
	\begin{myEnumerate}
		\item Choose a zipcode $s_c$ from possible zipcodes $Z$ and a neighborhood size $n$. The neighborhood of $n$ data-generating locations from $Z$ around $s_c$ is called $S_{cn}$.
		\item Choose a sparsity parameter $p \in (0,1]$. Choose locations from $S_{cn}$ with probability $p$ to form a subset $S$ of locations that will produce documents affected by the background and foreground topics. All other locations will produce documents generated from the background topics only.
		\item Choose a Bernoulli severity parameter $\gamma_i \in (0,1]$ for the $i^{th}$ zipcode in $S$ from $Beta(\eta)$.
		\item Choose a $T$-dimensional Dirichlet hyperparameter $\alpha$ for generating document-specific topic distributions.
		\item Choose a $W$-dimensional Dirichlet hyperparameter $\beta_f$ for generating foreground topics.
		\item Sample $T_f$ foreground topics from $Dir(\beta_f)$ together denoted by $\phi_f$; each topic $\phi_{fi}$ is a $W$-dimensional multinomial distribution over words.
		\item Denote the set of all topics as $\phi$ which includes topics from $\phi_b$ indexed from $1,..,T_b$ and topics from $\phi_f$ indexed from $T_b+1,..,T_b+T_f$
		\item For each of the $i=1,..,D_f$ foreground documents:
		\begin{myEnumerate}
			\item Sample a timestamp uniformly from the set of possible foreground timestamps $K_f$.
			\item Sample a zipcode $l_i$ uniformly from the set of zipcodes $Z$ considered to generate the documents.
			\item If $l_i \notin S$ or ($l_i \in S$ and $Bern(\gamma_{l_i}) == 0$),
			\begin{myEnumerate}
				\item Set new topic indicator $\delta_i=0$
				\item Sample a $T_b$-dimensional multinomial distribution $\theta_i$ over background topics from $Dir(\alpha_b)$.
				\item For each of the $j=1,..,N_i$ words in the document:
				\begin{myEnumerate}
					\item Sample a topic $z_{ij}$ for the word position $j$ from $Mult(\theta_i)$. $z_{ij}$ is one of the background topics $\phi_b$.
					\item Sample a word $w_{ij}$ for the word position $j$ from $Mult(\phi_{bz_{ij}})$.
				\end{myEnumerate}
			\end{myEnumerate}
			\item If $l_i \in S$ and $Bern(\gamma_{l_i}) == 1$,
			\begin{myEnumerate}
				\item Set new topic indicator $\delta_i=1$
				\item Sample a $T$-dimensional multinomial distribution $\theta_i$ over all topics from $Dir(\alpha)$.
				\item For each of the $j=1,..,N_i$ words in the document:
				\begin{myEnumerate}
					\item Sample a topic $z_{ij}$ for the word position $j$ from $Mult(\theta_i)$. $z_{ij}$ is one of the topics in $\phi_i$, which could be either a background or a foreground topic.
					\item Sample a word $w_{ij}$ for the word position $j$ from $Mult(\phi_{z_{ij}})$.
				\end{myEnumerate}
			\end{myEnumerate}
		\end{myEnumerate}
	\end{myEnumerate}
\end{myEnumerate}

\subsection{Inference for Spatially Compact Semantic Scan}
\label{sec:scss-inference}

We note that the variables marked in a hatched texture in the plate diagram of figure (\ref{fig:scss-plate-model}) are observed, while the other variables are to be inferred. The inference proceeds through MCMC sampling whose stationary distribution gives us the posterior distribution over the unobserved variables.

The inference of posterior distribution over variables $s_c$, $n$, $S_{cn}$, $p$, and $S$ (denoted collectively by $\mathcal{S}$) is done using the Generalized Fast Subset Sums framework (\cite{shao2011generalized}) which allows for efficient inference of these variables given the likelihood ratio $LR_i$ of each document.

This results in an alternating MCMC where the inference over $\mathcal{S}$ happens conditioned on the values over the remaining variables $\Omega-\mathcal{S}$, and the sampling of the variables $\Omega-\mathcal{S}$ proceeds conditioned on the inference for $\mathcal{S}$. We iterate between these two conditional inference steps until convergence.

\subsubsection{Inference over $\mathcal{S}$}

The likelihood ratio of $i^{th}$ foreground document $C_{fi}$ is given as

\begin{equation}
LR_i = \frac{\mathcal{L}(C_{fi} \mid \theta_i,~\phi)}{\mathcal{L}(C_{fi} \mid \theta_{bi},~\phi_b)}
\end{equation}

Since we do not know the exact parameters $\theta_i$, $\phi$, $\theta_{bi}$, and $\phi_b$, we use MAP estimates of these parameters from the collapsed Gibbs sampling phase of the inference.

We calculate the likelihood of a document with words $\mathbf{w}$ as follows:

\begin{align}
\mathcal{L}(\mathbf{w} \mid \hat{\theta},~\hat{\phi}) & = \sum_{\mathbf{z}} Pr(\mathbf{w},~\mathbf{z} \mid \hat{\theta},~\hat{\phi}) \\
& = \prod_{i=1}^N \left\{ \sum_{z_i} Pr(w_i, z_i \mid \hat{\phi}_{z_i}, \hat{\theta}) \right\} \\
& = \prod_{i=1}^N \left\{ \sum_{z_i} Mult(w_i \mid \hat{\phi}_{z_i}) \cdot Mult(z_i \mid \hat{\theta}) \right\}
\end{align}

Here, the last step of interchanging sum of products with product of sums can be carried out because we are marginalizing over all possible values of each topic assignment variable $z_i$. This considerably speeds up the computation of the likelihood terms from exponential to linear time.

Once we have calculated the likelihood ratio for each document $LR_i$, we can calculate the posterior probability of the event $E_{cn}$ that emerging topics are localized in a given neighborhood $S_{cn}$ centered at $s_c$ and consisting of $n$ locations as follows (\cite{shao2011generalized}):-

\begin{align}
Pr(E_{cn} \mid C_f) & \propto \sum_{S \subseteq S_{cn}} Pr(S \mid C_f) \\
& \propto \sum_{S \subseteq S_{cn}} Pr(S) \cdot \prod_{s_i \in S} LR_i \\
& \propto \sum_{S \subseteq S_{cn}} p^{|S|} \cdot (1-p)^{n-|S|} \cdot \prod_{s_i \in S} LR_i \\
& \propto (1-p)^n \sum_{S \subseteq S_{cn}} \left(\frac{p}{1-p}\right)^{|S|} \cdot \prod_{s_i \in S} LR_i \\
& \propto (1-p)^n \sum_{S \subseteq S_{cn}} \prod_{s_i \in S} \left(\frac{p}{1-p}\right) \cdot LR_i
\end{align}

Since we are summing over $2^n$ subsets of $S_{cn}$, we can reduce the time complexity from exponential to linear by writing sum of $2^n$ products as the product of $n$ sums (\cite{shao2011generalized}).

\begin{align}
\therefore Pr(E_{cn} \mid C_f) & \propto (1-p)^n \prod_{s_i \in S_{cn}} \left\{ 1 + \left(\frac{p}{1-p}\right) \cdot LR_i \right\} \\
& \propto \prod_{s_i \in S_{cn}} \left\{ 1-p + p \cdot LR_i \right\}
\end{align}

Thus, the posterior probability of a neighborhood $S_{cn}$ showing an outbreak of foreground topics in its documents is proportional to the product of smoothed likelihood ratios $(1-p + p \cdot LR_i)$ for all documents in the neighborhood. Finally, we calculate the normalizer by marginalizing over all $S_{cn}$.

\begin{align}
Pr(E_{cn} \mid C_f) & = \frac{Pr(E_{cn} \mid C_f)}{\sum_{\forall~s_c,n} Pr(E_{cn} \mid C_f)}
\end{align}

Probability of an event $E_j$ that foreground topics occur in document at location $s_j$ can be calculated in a similar fashion by considering only those neighborhoods $S_{cn}$ where $s_j$ is included and summing over the $2^{n-1}$ subsets $S \subseteq S_{cn}$ such that $s_j \in S$. The average likelihood ratio in neighborhood $S_{cn}$ such that location $s_j$ is always included in any chosen subset $S$ of locations is given as $p \cdot LR_j \cdot \prod_{s_i \in S_{cn}-s_j} \left\{ 1-p + p \cdot LR_i \right\}$. The total posterior probability $Pr(E_j \mid C_f)$ of document in location $s_j$ containing the foreground topics can be calculated by marginalizing over all $s_c$ and $n$ such that the resultant neighborhood $S_{cn}$ contains location $s_j$.

\subsubsection{Inference over $\Omega - \mathcal{S}$}

This phase is very similar to the second phase of Semantic Scan inference described earlier with changes to incorporate only those foreground documents that we believe actually contain the emerging topics and are therefore composed of not just the background topics. Given the total posterior probability $Pr(E_i \mid C_f)$ of document in location $s_i$ containing the foreground topics calculated from the previous phase of inference, we generate a binary value $\delta_i$ by sampling from a Bernoulli distribution $Bern(Pr(E_i \mid C_f))$. If $\delta_i=0$, we do not believe that the document has foreground topics and therfore do not include it as a part of the foreground documents on which Gibbs sampling happens in the second phase of Semantic Scan. If $\delta_i=1$, we believe that the document has foreground topics as indicated by our spatial inference based on document-specific likelihood ratios and therfore we include the document as a part of the foreground documents on which Gibbs sampling happens in the second phase of Semantic Scan. Once we have decided the documents on which Gibbs sampling is to be done by assigning values to $\delta_i$ , we proceed with collapsed Gibbs sampling according to the procedure described in section (\ref{sec:ss-inference}) to obtain foreground topics $\phi_f$ and document-specific distributions over topics $\theta_i$ that are used again in inference over $\mathcal{S}$ in our alternating inference mechanism.

\begin{figure}
\begin{center}
    \begin{tabular}{c  c}
    \includegraphics[page=3,trim=20mm 20mm 20mm 20mm,clip=true,width=0.45\columnwidth]{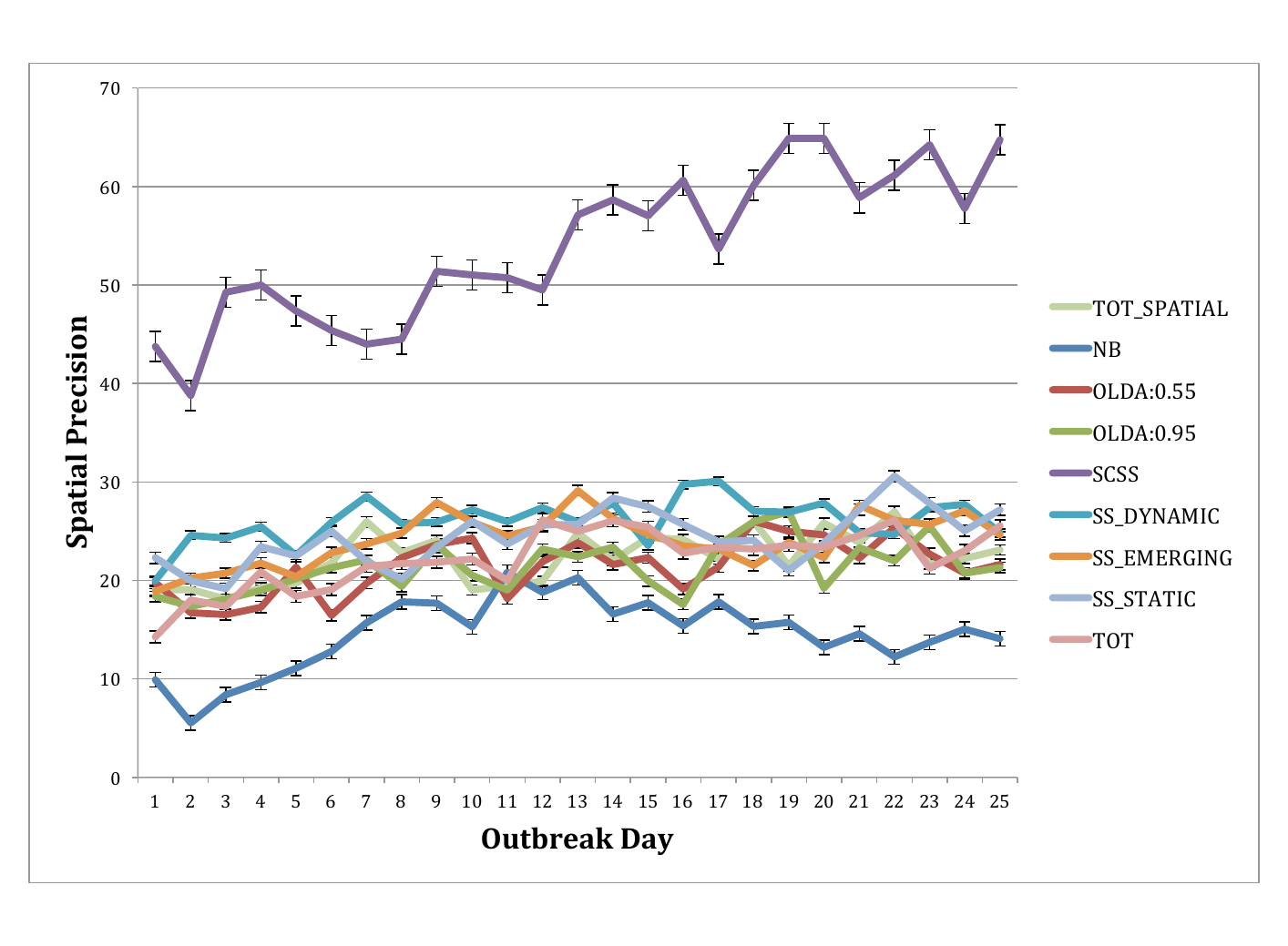} & \includegraphics[page=6,trim=20mm 20mm 20mm 20mm,clip=true,width=0.45\columnwidth]{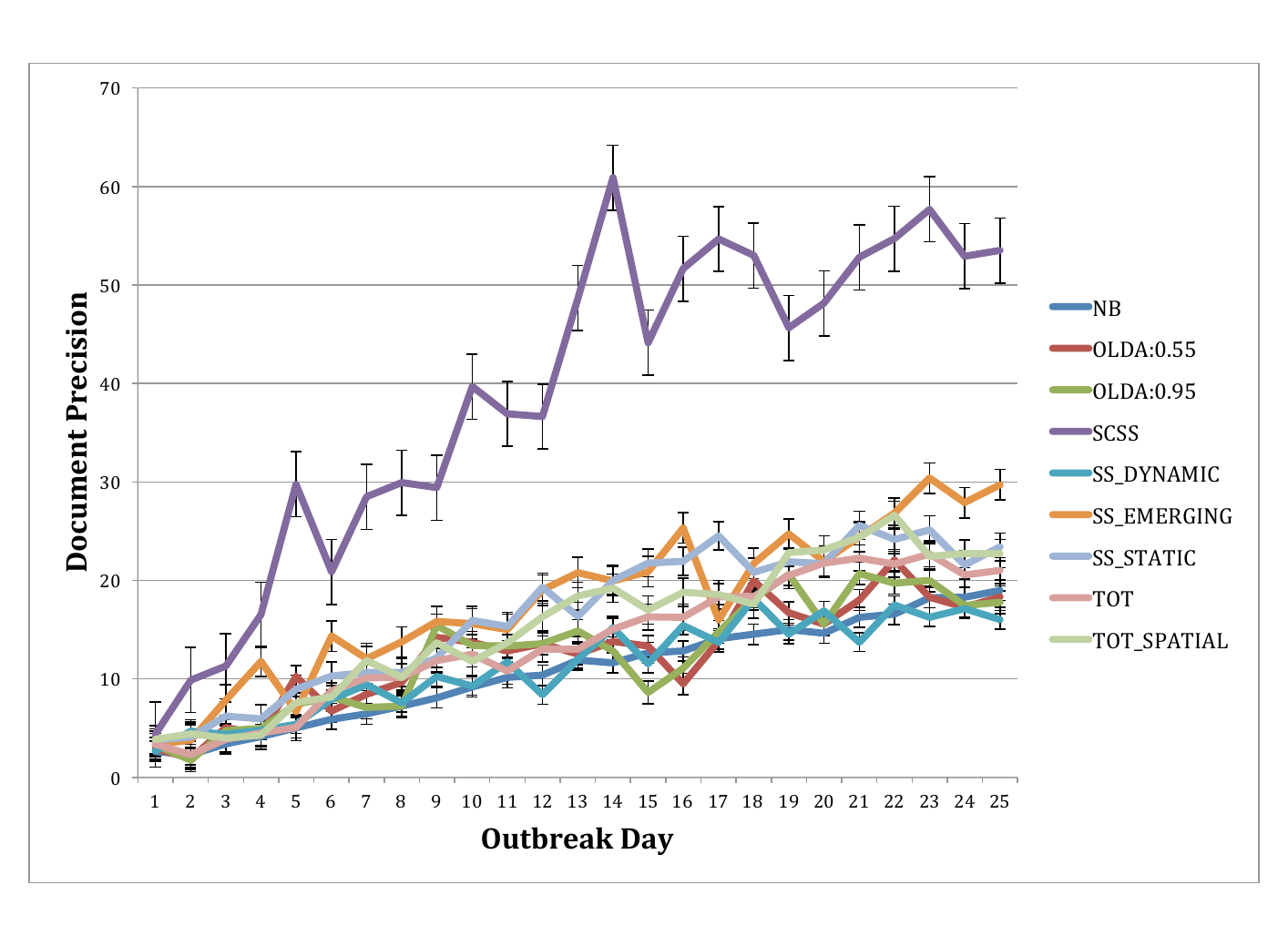} \\
	(a) Spatial Precision & (a) Document Precision
    \end{tabular}
\end{center}
\label{fig:precision}
\end{figure}
\begin{figure}
\begin{center}
    \begin{tabular}{c  c}
    \includegraphics[page=4,trim=20mm 20mm 20mm 20mm,clip=true,width=0.45\columnwidth]{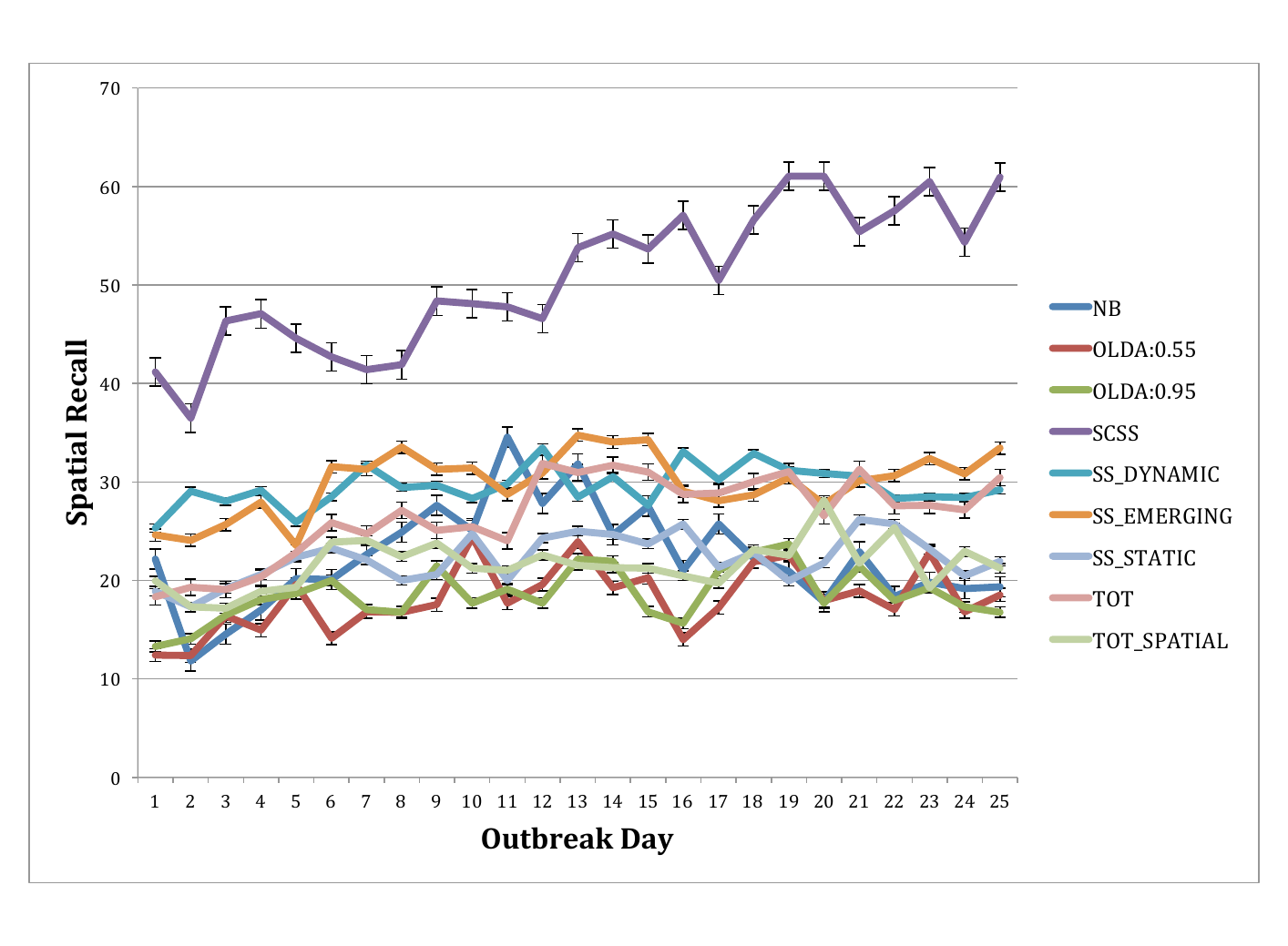} & \includegraphics[page=7,trim=20mm 20mm 20mm 20mm,clip=true,width=0.45\columnwidth]{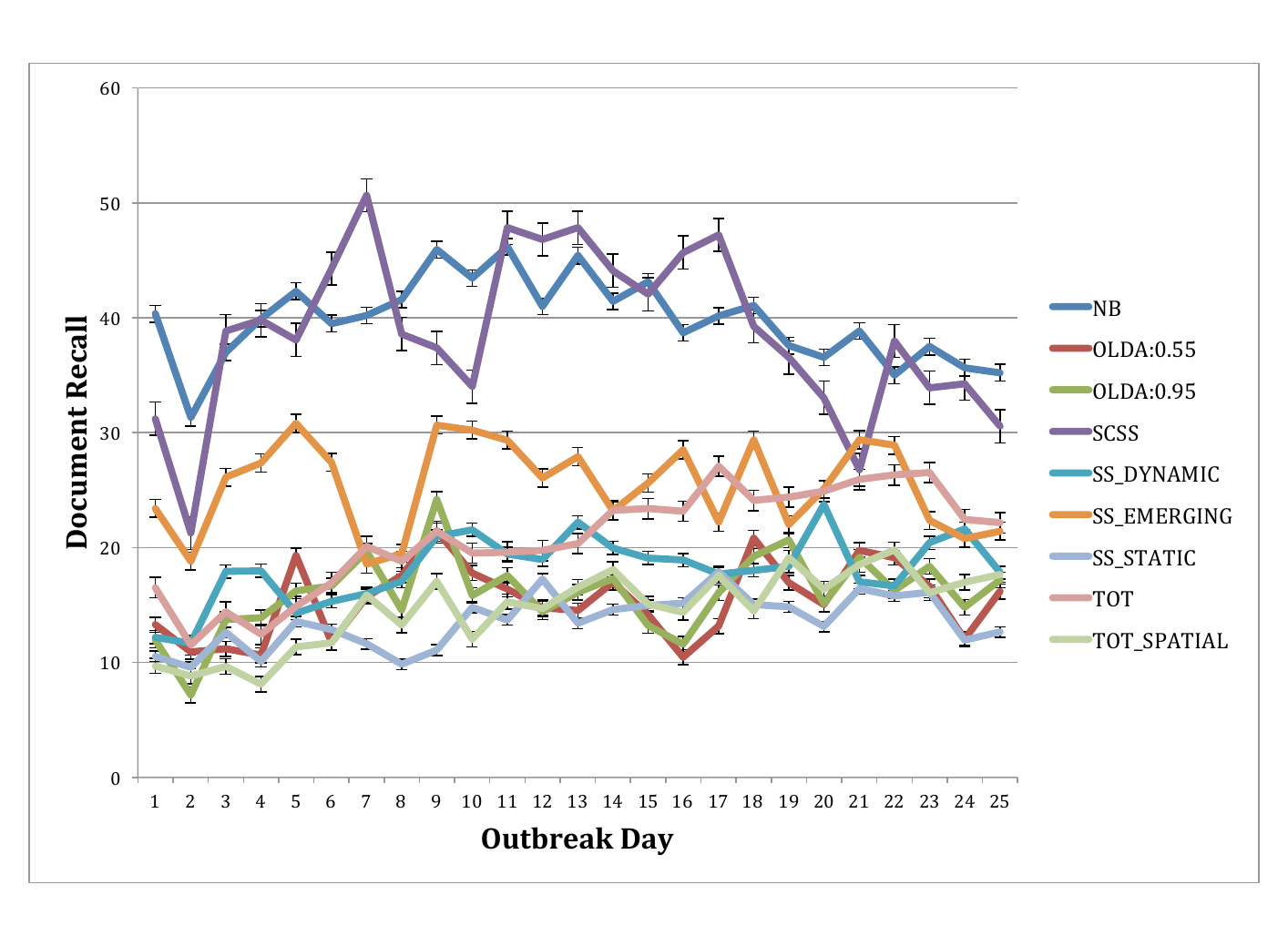} \\
	(b) Spatial Recall & (b) Document Recall
    \end{tabular}
\end{center}
\label{fig:recall}
\end{figure}
\begin{figure}
\begin{center}
    \begin{tabular}{c  c}
    \includegraphics[page=5,trim=20mm 20mm 20mm 20mm,clip=true,width=0.45\columnwidth]{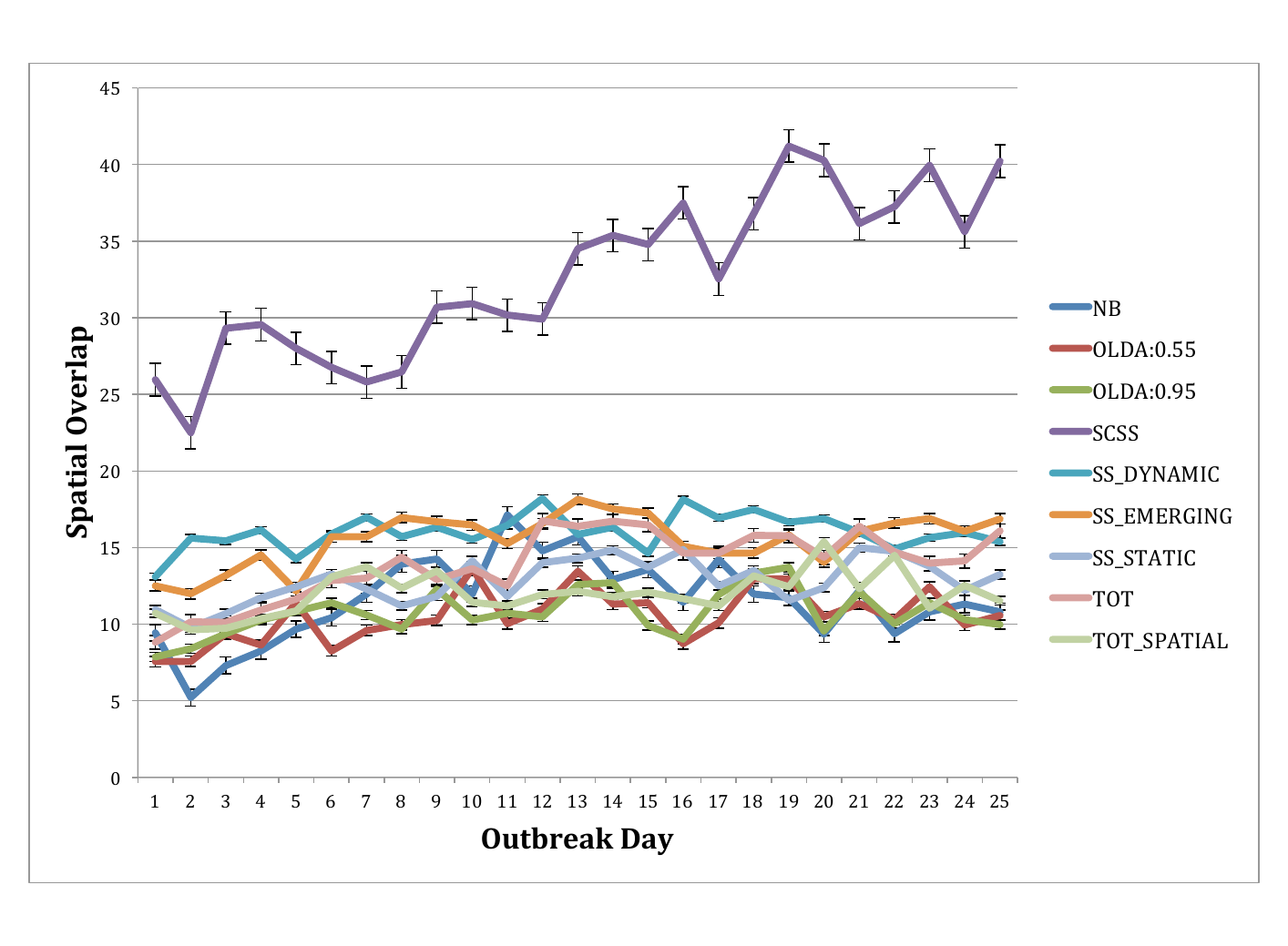} & \includegraphics[page=8,trim=20mm 20mm 20mm 20mm,clip=true,width=0.45\columnwidth]{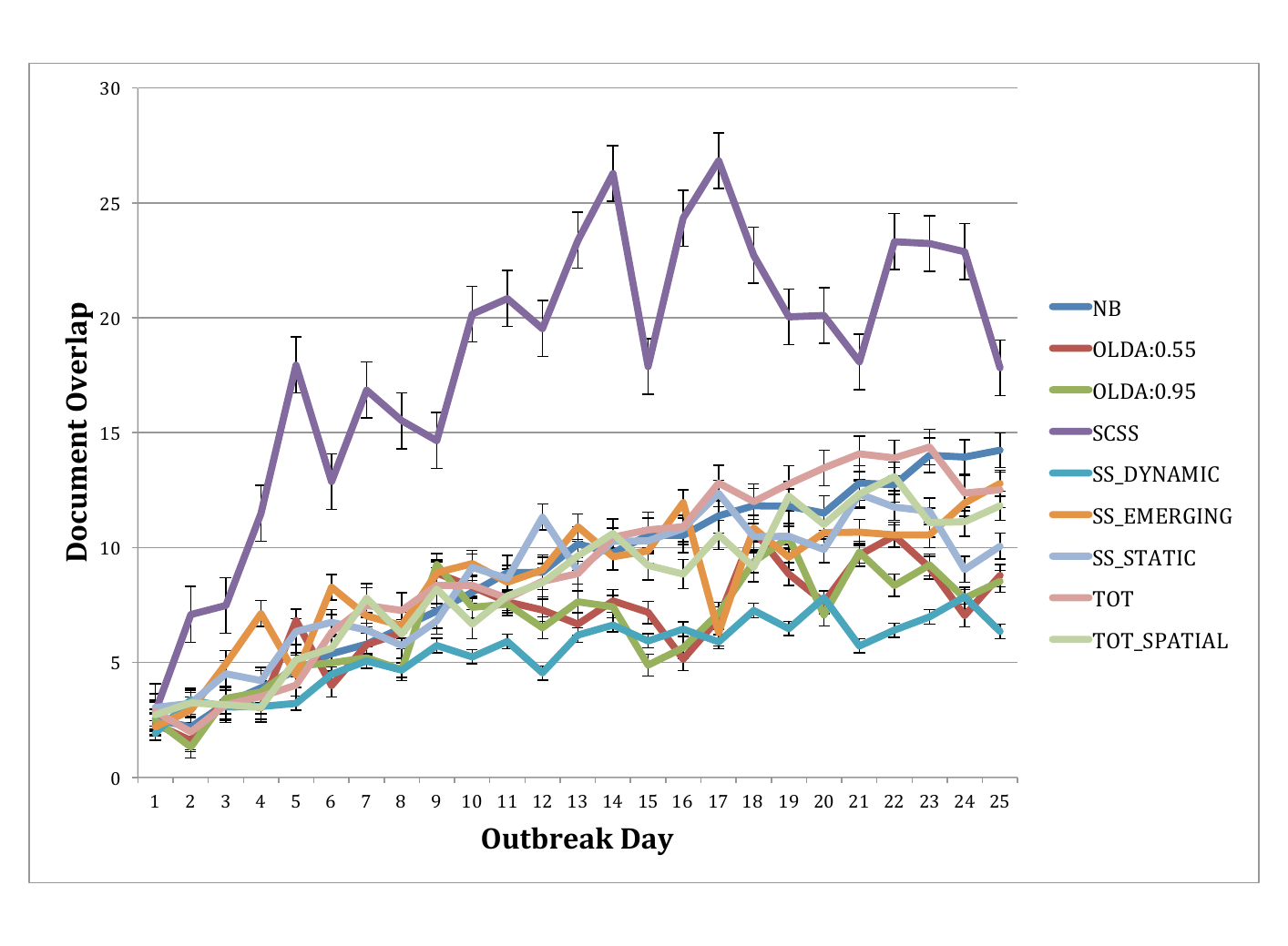} \\
	(c) Spatial Overlap & (c) Document Overlap
    \end{tabular}
\end{center}
\label{fig:overlap}
\end{figure}

\section{Results}
\label{sec:results}

In this section, we describe the ED dataset and how we use it to compare Spatially Compact Semantic Scan (SCSS) with competing approaches such as Semantic Scan (SS), Topics over Time (ToT), Spatial Topics over Time (Spatial ToT), Online LDA (OLDA), and Naive Bayes (NB).

\subsection{Emergency Department Chief Complaints Dataset}

The ED dataset consists of text complaints noted by the staff of emergency departments of Allegheny County hospitals. The dataset includes complaints from 2003 to 2005. Each complaint is associated with the date on which it was recorded, the zipcode of the hospital where the complaint was recorded, and the ICD9 code to which it was assigned. The external manual classification of diseases using the ICD9 codes has been used to create semi-synthetic disease outbreaks as described below. This external piece of information associated with a text complaint is not assumed to be known by the detection methods and is used for evaluating the methods only. In practice, in many cases, the ICD9 code is unknown or incorrect until its final assignment for billing purposes after the patient's visit. In order to get geospatial coordinates for a complaint, we map its associated zipcode to the centroid latitude and longitude coordinates for the zipcode area. Thus, we have a dataset where each document is a short text complaint followed by the date on which it was recorded, the geospatial coordinates of the zipcode in which it was recorded, and its ICD9 code.

\begin{figure}
\begin{center}
    \begin{tabular}{c  c}
    \includegraphics[page=1,trim=20mm 20mm 20mm 20mm,clip=true,width=0.5\columnwidth]{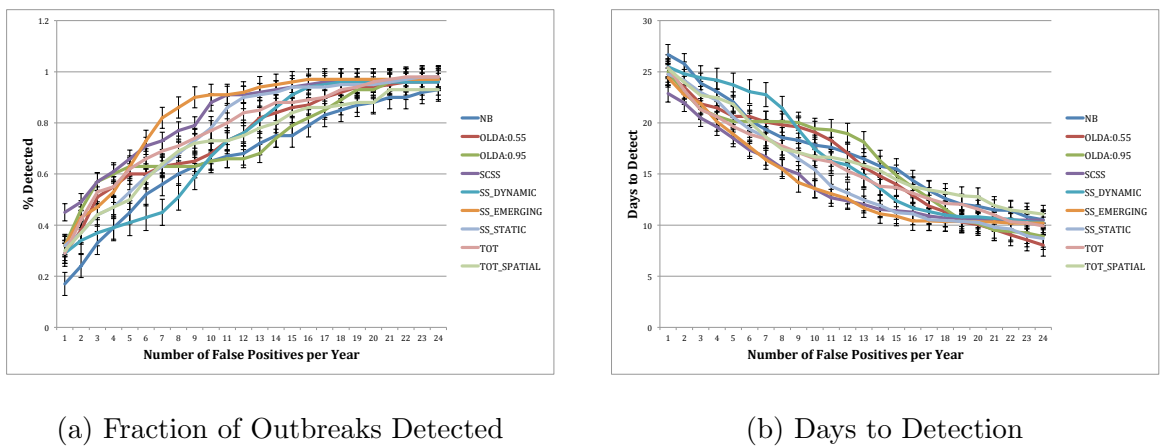} & \includegraphics[page=2,trim=20mm 20mm 20mm 20mm,clip=true,width=0.5\columnwidth]{images/graphs} \\
	(a) Fraction of Outbreaks Detected & (b) Days to Detection \\
    \end{tabular}
\end{center}
\caption{Detection Power}
\label{fig:detection-power}
\end{figure}

\subsection{Experimental Setup}

We perform leave-one-out (LOO) validation of SCSS alongwith the baseline methods. We treat documents from 2003 as the background documents and documents from 2004 as the foreground documents. We pick the 10 most frequent ICD9 codes in the dataset. For each of these ICD9 codes, we remove all complaints from the background and foreground data corresponding to that ICD9 code. We then create outbreaks corresponding to the held-out ICD9 code in the foreground documents belonging to 2004. The outbreak is created by sampling $s_c$ and $n$, calculating the neighborhood $S_{cn}$, sampling sparsity parameter $p$, and sampling $S$ from $S_{cn}$. To generate a datapoint in an outbreak, we sample the text document uniformly from the held-out ICD9 text complaints, and the location uniformly from the zipcodes in $S$. For each of the 10 held-out ICD9 codes, we create 10 outbreaks each, resulting in a total of 100 outbreaks over which we run SCSS and each of the baselines. Each outbreak is 30-days long, and the number of cases generated for the $d^{th}$ day is $3*d$. While running any of the methods, we assume a 3-day moving window for outbreak detection. For methods like SCSS and Spatial ToT, it is not necessary to perform a spatial scan as the last step since these methods also output a detected spatial region. For other methods including the original semantic scan, we perform the assignment of documents to topics and circular spatial scan as outlined in the Semantic Scan paper (\cite{liu2011detecting}).

\subsection{Competing Approaches}

We have chosen the following related work against which to compare SCSS:

\begin{itemize}
	\item \textbf{SS-Emerging:} This is the version of Semantic Scan (SS) (\cite{liu2011detecting}) that we described in section (\ref{sec:methodology}). It allows for both background and foreground topics and holds background topics fixed while learning foregroudn emerging topics on incoming batch of documents. 25 background and 25 foreground topics are used in the evaluation.
	\item \textbf{SS-Dynamic:} This version of SS (\cite{liu2011detecting}) does not allow for any background topics. The topic learning only focuses on foreground topics on an incoming batch of documents. 25 foreground topics are used in the evaluation; there are no background topics in this model. 
	\item \textbf{SS-Static:} This version of SS (\cite{liu2011detecting}) does not allow any foreground topics. Once static topics are learned at the beginning, SS-Static only performs document assignment and sptial scan steps for an incoming batch of documents. 25 background topics are used in the evaluation; there are no foreground topics in this model. 
	\item \textbf{Topics over Time (ToT):} ToT (\cite{wang2006topics}) is an LDA variant that incorporates temporal aspect of a document set by modeling the timestamps assigned to a topic as being sampled from a $Beta$ distribution specific to the topic. 50 total topics are used in the evaluation of this this model. 
	\item \textbf{Spatial Topics over Time (Spatial ToT):} In order to perform an apples-to-apples comparison to SCSS, we modified ToT (\cite{wang2006topics}) by additionally modeling the spatial coordinates assigned to a topic as being sampled from a 2D spatial $Gaussian$ distribution specific to the topic. 50 total topics are used in the evaluation of this this model.
	\item \textbf{Online LDA:} We compare SCSS to two versions of Online LDA (\cite{hoffman2010online}) - one with $\kappa=0.55$, and one with $\kappa=0.95$. We refer to these two variants as "OLDA:0.55" and "OLDA:0.95" respectively. $\kappa$ is a hyperparameter of Online LDA algorithm that controls how quickly topics can adapt to changes in the topics of the text stream. 50 total topics are used in the evaluation of this this model.
	\item \textbf{Naive Bayes:} Finally, we compare SCSS to NB by considering background and foreground documents as belonging to two different classes, using the NB prediction as a document's assignment, and performing spatial scan on the obtained document assignments.
\end{itemize}

\subsection{ED Dataset Results}

We consider the following metrics to compare SCSS to our chosen baselines:

\begin{itemize}
	\item \textbf{Spatial Precision:} $\left(\frac{tp}{tp+fp}\right)$ The fraction of zipcodes that are actually a part of the outbreak, out of the zipcodes that were detected to be a part of the outbreak.
	\item \textbf{Spatial Recall:} $\left(\frac{tp}{tp+fn}\right)$ The fraction of zipcodes that were detected to be a part of the outbreak, out of the zipcodes that are actually a part of the outbreak.
	\item \textbf{Spatial Overlap:} $\left(\frac{tp}{tp+fp+fn}\right)$  The fraction of zipcodes that are actually a part of the outbreak as well as detected to be a part of the outbreak, out of all zipcodes that were either actually a part of the outbreak or detected to be a part of the outbreak.
	\item \textbf{Document Precision:} $\left(\frac{tp}{tp+fp}\right)$ The fraction of foreground documents that actually contain the foreground topics out of the foreground documents that were detected to contain the foreground topics.
	\item \textbf{Document Recall:} $\left(\frac{tp}{tp+fn}\right)$ The fraction of foreground documents that were detected to contain the foreground topics out of the foreground documents that actually included the foreground topics during their generation.
	\item \textbf{Document Overlap:} $\left(\frac{tp}{tp+fp+fn}\right)$  The fraction of documents that were injected with foreground topics and detected to contain the foreground topics out of all documents that were either injected with foreground topics or detected to contain the foreground topics.
	\item \textbf{Percentage of Outbreaks Detected:} The percentage of outbreaks detected versus the number of false positives per year. This is a monotonically non-decreasing graph where a higher value represents a better outcome.
	\item \textbf{Days to Detection:} The number of days of data required to detect an outbreak versus the number of false positives per year. This is a monotonically non-increasing graph, where a lower value represents a better outcome.
\end{itemize}

The graphs for spatial precision, recall, and overlap can be found in figure (\ref{fig:spatial-metrics}). The three metrics are plotted against the outbreak day on the X-axis which ranges from 1 to 25. We observe that SCSS has spatial precision, recall, and overlap almost double that of any of the baselines. All three metrics improve steadily as the duration and intensity of the outbreak increases.

The graphs for document precision, recall, and overlap can be found in figure (\ref{fig:document-metrics}). The three metrics are again plotted against the duration of the outbreak on the X-axis measured in number of days. We notice that SCSS has significantly better document precision and overlap compared to the baselines. For document recall, SCSS and Naive Bayes have similar performance, and Naive Bayes exceeds the SCSS performance at several points of the graph. However, this just indicates that Naive Bayes is classifying a lot of documents as a part of the outbreak. Considered together with document precision and overlap, we still conclude that SCSS performs significantly better than the baselines we have compared to.

The graphs for the fraction of outbreaks detected and the number of days of data required to detect an outbreak can be found in figure (\ref{fig:detection-power}). Both metrics are plotted against the number of false positives per year on the X-axis. As expected, we see that the fraction of outbreaks detected increases as we allow more false positives per year. Similarly, the number of days of data required to detect an outbreak decreases as we allow more false positives per year. We note that SCSS performance on these metrics is comparable to that of SS-Emerging, and is better than SS-Emerging for low false positive rates. SCSS performance is not significantly better than the baselines on these two metrics. However, coupled with performance on precision, recall, and overlap metrics, SCSS beats the baselines that we have compared to. Many of these baselines like ToT, Spatial ToT (which performs better than ToT), and Online LDA are state-of-the-art methods in literature for (spatio-)temporal event detection in text streams.

\section{Future Work}
\label{sec:future-work}

We envision the following possible investigations and refinements to SCSS:

\begin{itemize}
	\item Combining spatio-temporal text data streams with other spatio-temporal data such as heat indices, medication sales, etc. to improve the detection power of SCSS.
	\item Testing the robustness of SCSS using other datasets and outbreak simulations.
	\item Scaling up the method so that it can be run on massive datasets such as Yelp reviews or Twitter streams.
	\item Ability to detect multiple spatial clusters. Prior work (\cite{zhang2010spatial}) incrementally detect a cluster and removes it to reveal other clusters. Incorporating this feature and testing its accuracy and efficiency merit further investigation.
\end{itemize}

\section{Conclusions}
\label{sec:conclusions}

We have proposed a topic model for finding spatially compact and temporally emerging topics in real-world text corpora. We have evaluated the model on real-world ED data from disease outbreak detection and presented our results in section (\ref{sec:results}) to demonstrate the efficacy of our method.

One of the promising future directions we are considering is finding subtle emerging topics by mining the residuals of the new documents i.e. the component of the document vectors not explained by the currently learnt topics. A newly emerging topic will tend to create clusters in the residual space, which can then be mined for topics in high density regions using an algorithm like DBSCAN (\cite{ester1996density}), a spatial data-structure like R-Tree (\cite{guttman1984r}), or using linear algebraic scan statistics that search for high density cones in the residual space.

\newpage

\bibliography{report}
\bibliographystyle{unsrt}

\end{document}